\documentclass{article}

     \PassOptionsToPackage{numbers, compress}{natbib}
     \usepackage[preprint]{neurips_2021}

\usepackage[utf8]{inputenc} % allow utf-8 input
\usepackage[T1]{fontenc}    % use 8-bit T1 fonts
\usepackage{hyperref}       % hyperlinks
\usepackage{url}            % simple URL typesetting
\usepackage{booktabs}       % professional-quality tables
\usepackage{amsfonts}       % blackboard math symbols
\usepackage{nicefrac}       % compact symbols for 1/2, etc.
\usepackage{microtype}      % microtypography
\usepackage{xcolor}         % colors
\usepackage[pdftex]{graphicx}
\usepackage{amsmath}
\usepackage{caption}
\usepackage{subcaption}

\title{Embed Everything: A Method for Efficiently Co-Embedding Multi-Modal Spaces}

\author{%
  Sarah Di\thanks{Equal contribution.}\:\:\thanks{Work performed while at Soot.} \\
  Department of Computer Science\\
  Carnegie-Mellon University\\
  Pittsburgh, PA 15213 \\
  \texttt{sarahdi@andrew.cmu.edu} \\
  \And
  Robin Yu\footnotemark[1]\:\:\footnotemark[2] \\
  Department of Computer Science\\
  Harvey Mudd College\\
  Claremont, CA 91711 \\
  \texttt{rhyu@g.hmc.edu} \\
  \And
  Amol Kapoor\\
  Soot\\
  Brooklyn, NY, 11216\\
  \texttt{amol@soot.com} \\
}

\begin{document}

\maketitle

\begin{abstract}
  Any general artificial intelligence system must be able to interpret, operate on, and produce data in a multi-modal latent space that can represent  audio, imagery, text, and more. In the last decade, deep neural networks have seen remarkable success in unimodal data distributions, while transfer learning techniques have seen a massive expansion of model reuse across related domains. However, training multi-modal networks from scratch remains expensive and illusive, while heterogeneous transfer learning (HTL) techniques remain relatively underdeveloped. In this paper, we propose a novel and cost-effective HTL strategy for co-embedding multi-modal spaces. Our method avoids cost inefficiencies by preprocessing embeddings using pretrained models for all components, without passing gradients through these models. We prove the use of this system in a joint image-audio embedding task. Our method has wide-reaching applications, as successfully bridging the gap between different latent spaces could provide a framework for the promised "universal" embedding.
\end{abstract}

\section{Introduction}

Traditional deep neural network classifiers (ImageNet, Inception, etc.) \cite{imagenet,inception,unimodal1}  operate in a unimodal latent space. Unimodal spaces, though useful, fail to encapsulate the inherently multi-modal methods that human beings use to interact with and experience real-world data. Bridging modalities is an active area of research \cite{multimodalml,zhang2017multi}, frequently through a combination of transfer learning and the use of deep neural networks \cite{mithun2019joint}.

Although deep neural networks exhibit remarkable generalization capabilities \cite{unreasonable}, even the best neural networks perform poorly on data that is far from the training distribution. A naive approach is to retrain models on unique sub-tasks; however, training costs increase in proportion to model size and complexity, and many tasks do not have sufficient labeled data available to fully train a model from scratch. 

Transfer learning has proven to be an effective tool in generalizing knowledge across domains with similar feature spaces \cite{transferlearning}. Transfer learning can involve a model architecture that utilizes some or all of an existing trained model to generate learned features. This model is then trained on a (traditionally much smaller) dataset for a specific subtask. Transfer learning has successfully been used for text sentiment classification, image classification, and multi-language text classification \cite{weiss2016survey}. Notably, in many of these cases, the source and target domains involve similar feature spaces. In instances where the source and target domain are highly dissimilar (e.g. across modalities), standard transfer learning methods fail.

Heterogeneous transfer learning (HTL), which aims to connect nonequivalent feature spaces, requires complex embedding transformations in order to handle cross-domain differences in data. HTL techniques can significantly broaden real-world applications of transfer learning -- tasks as varied as search or autonomous vehicle training depend on multi-modal inputs, including audio, imagery, and text\cite{multimodalml}. Ideally, HTL techniques would allow practitioners to compose existing models that perform well in specific modalities, resulting in complex co-embedding spaces that function \textit{across} modalities.

In this work, we aim to tackle this foundational problem in HTL. We propose a heterogeneous transfer learning method (Embed Everything) which can be used to learn a co-embedding space over any two arbitrary pretrained models. Our approach utilizes preprocessed embeddings generated by the pretrained models without passing gradients to the models themselves, thereby distributing computational work while minimizing cost. We deploy contrastive losses to train a small deep neural network that projects between the preexisting model spaces. Using Embed Everything, we develop an image-audio co-embedding space that is based on CLIP \cite{clip} and VGGISH \cite{vggish}. We present several experiments to show the efficacy of the Embed Everything method, and conclude with an analysis of the applications and limitations of our approach.

\section{Related Work}
\label{gen_inst}

\subsection{Transfer Learning} Transfer learning techniques aim to make model training more efficient by reusing parts or all of previously trained models for feature extraction. Models trained with transfer learning generally require fewer training steps \cite{transferlearning, heterogeneoustl,friedjungova2017asymmetric} and smaller datasets \cite{transferlearning,heterogeneoustl,friedjungova2017asymmetric} to generalize to new tasks. Deep neural networks work well with transfer learning techniques because lower layers of the model learn successively more abstract features of the dataset, which are more easily transferred to unrelated tasks \cite{bengio2012deep}. 

Transfer learning on deep neural networks can be described with the following notation:
\begin{align*}
    h_0&=x\\
    h_1&=f_1(h_0)\\
    h_n&=f_n(h_{n-1})\\
    t_0&=h_{1...n}\\
    t_1&=g_1(t_0)\\
    t_n&=g_n(t_{n-1})\\
\end{align*}
where $x$ represents input data, $f_{1...n}$ represent pre-trained learnable functions that are composed as layers in a deep neural network, $g_{1...n}$ represents the same but randomly initialized, $h_{1...n}$ represents the embeddings from a pretrained model $H$ at different layers, and $t_{1...n}$ represents the embeddings from a transfer-learned model $T$ at different layers. In the traditional setting, $T$ is then trained on new data. Gradients can optionally flow from $t_0$ to $h_{1...n}$; in such settings, the entire model is able to fine tune on the novel inputs, resulting in higher quality outputs but with greater training cost. 

The method of transfer learning described above relies on $T$ being trained on a task that is fundamentally similar to $H$, as both models depend on the same input data $x$. As a result, transfer learning techniques will not succeed on markedly different input data. Heterogeneous Transfer Learning techniques attempt to resolve this issue by training models that can explicitly project information into different target domains. We recommend \cite{heterogeneoustl} for an in depth review of existing HTL methods. As far as we are aware, HTL methods have not been applied to learning joint embedding spaces in multiple modalities, such as images and audio. 

\subsection{Multi-Modal Deep Learning} 

Deep neural networks have been shown to have powerful representation capabilities in multiple domains, including imagery, audio, and text. Because neural networks operate on vector space representations, they have become popular for attempting to solve multi-modal tasks \cite{multimodalml}. Specific architectures are often used in conjunction with individual modalities -- for example, convolutional networks for images, or transformers for text. As a result, multi-modal models are often composed of unimodal submodels, with composition layers that aggregate information into a single coherent representation that is then used to solve some underlying, potentially supervised task. 

More recently, models such as CLIP \cite{clip} have successfully created joint embedding spaces across modalities. Such models are extremely powerful, with CLIP displaying impressive generalization capabilities across a variety of tasks\cite{clip}. However, they are also expensive to train and require a significant amount of data. CLIP and similar models were trained using contrastive losses, an increasingly popular method of learning embedding spaces without supervision \cite{contrastivelearning}. In particular, contrastive losses have shown surprising ability to adapt to potentially unclean labeled data. We suspect that contrastive losses can be utilized to find correlations across preexisting multi-modal embedding spaces.

\section{Methods}

The core insight behind the Embed Everything method is that preexisting embedding spaces can be projected into arbitrary target domains using a relatively small neural network that can be fine-tuned for specific tasks. Our system uses the following components: two pretrained models, $M_1$ and $M_2$ that ingest different modalities $m_1$ and $m_2$ respectively; a set of untrained transform layers $T_1$ and $T_2$ on top of one or both models that are significantly smaller in size than the pretrained models that produce outputs of the same dimensionality; and a contrastive loss function on an optimizer that takes in inputs from $T_1$ and $T_2$. The Embed Everything method additionally requires a dataset of pairs between $m_1$ and $m_2$; such data-pairs are commonly found from scrapes of the web (e.g. in \cite{clip}, \cite{brown2020language}).

Training procedure is as follows:
\begin{enumerate}
    \item batches are created following the method laid out in \cite{contrastivelearning} for the contrastive loss;
    \item batches are fed into $M1$ and $M2$ to produce embeddings in their own custom spaces, potentially of different dimensionalities;
    \item these custom embeddings are passed through the transform layers $T1$ and $T2$, producing an embedding of the same size;
    \item the same-sized embeddings are passed to the contrastive loss, which produces gradients such that embeddings coming from a pair are similar while all others are distinct;
    \item gradients are passed down through the model, which results in the transform layers learning to project into a common space.
\end{enumerate}

We could equivalently denote the outputs as follows: 
\begin{align*}
    e_1 = T_1(M_1(m1))\\
    e_2 = T_2(M_2(m2))\\
\end{align*}
where $e_1$ and $e_2$ denote embeddings that live in the same latent space for potentially multimodal inputs $m1$ and $m2$.

During inference, data is passed into $M1$ and $M2$ as usual. Embeddings taken from the pretrained models are then fed into the now-learned and static transform layers. Because the transform layers have been trained to produce embeddings that live in the same vector space, the outputs should be comparable, as if they were produced by the same model. 

To massively speed up training, we avoid passing gradients to $M1$ and $M2$. This allows us to separate steps 1 and 2 from the rest of the training process described above. We can effectively treat $M1$ and $M2$ as black boxes and run them in inference mode at scale in a highly distributed setting. We can then work entirely embedding spaces, discarding any of the original data. The entire model can now be seen as training a few small projection layers on input float vectors, which is a significantly less expensive task than training huge models from high dimensional data like images and audio.

\begin{figure}
  \centering
  \includegraphics[width=\linewidth]{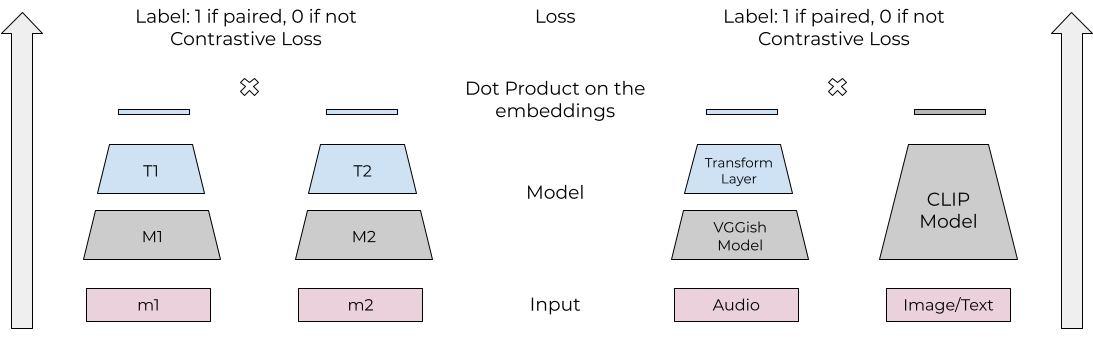}
  \caption{\textbf{Summary of our approach}. {\it Left:} Our system comprises two pretrained models, $M_1$ and $M_2$ which take as input $m_1$ and $m_2$ respectively and two untrained transform layers $T_1$ and $T_2$. {\it Right:} First, we preprocess our training data by generating the embeddings produced by CLIP and VGGish on VGG-Sound videos. We train additional layers on top of the fixed audio embedding model to transform the VGGish embedding space to align with the CLIP embedding space.}
  \label{modelfig}
\end{figure}

\section{Experiments}
\label{headings}

In this section, we describe how Embed Everything was used to create a joint image-audio embedding space.

\subsection{Dataset}
\label{dataset}
In order to utilize Embed Everything, we need to have a relatively large dataset of image-audio pairs. Conveniently, we can utilize the large amount of video data available online to generate these pairs. We scrape X videos from the VGG-Sound dataset \cite{vggsound}. From each video, we extract image frames that correspond to five seconds of audio around the frame. This process generates X image-audio pairs. We randomly divide image-audio pairings 67\%/33\% to form the training and validation datasets.

\subsection{Model Settings}

For our experiments, we utilize CLIP for our pretrained image embedding space and VGGish for our pretrained audio embedding space. To further simplify our training scheme, we utilize a single transform layer that projects the VGGish output embeddings into the CLIP embedding space. Because we do not pass gradients to CLIP or VGGish, we preprocess all of the image-audio pairs described in \ref{dataset} offline, at scale. These embeddings are then turned into batches of data and labeled as positive or negative based on whether the image-audio selection represents a pair in the original dataset.

The transform layers are implemented in Keras and trained with an Adam optimizer using an adjusted learning rate of 0.0001. We train at batch size 4096 for 300 epochs. The transform layers were trained entirely on a personal laptop on CPU, underscoring the efficiency of the method. See Figure \ref{modelfig} for a visual description of the final approach.

\begin{figure}
     \centering
     \begin{subfigure}[b]{0.45\textwidth}
         \centering
         \includegraphics[width=\textwidth]{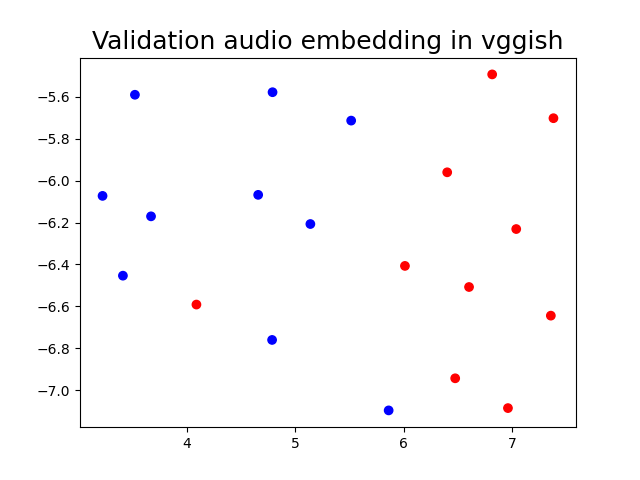}
     \end{subfigure}
     \begin{subfigure}[b]{0.45\textwidth}
         \centering
         \includegraphics[width=\textwidth]{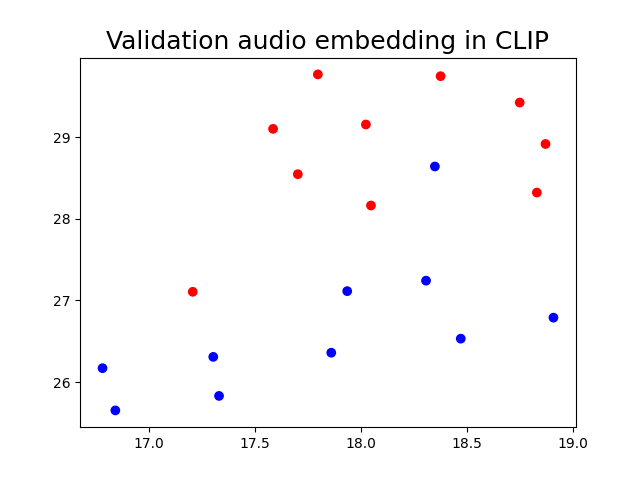}
     \end{subfigure}
     \caption{\textbf{Preservation of distinct classes}. UMAP visualization of embeddings of audio from two distinct classes. Audio from bells (blue) and cars (red) is embedded first in the VGGish space and then transformed into the CLIP space. This transformation retains separation of the two classes.}
    \label{exp1a}
    \label{exp1b}
     \centering
         \begin{subfigure}[b]{0.6\textwidth}
         \centering
         \includegraphics[width=\textwidth]{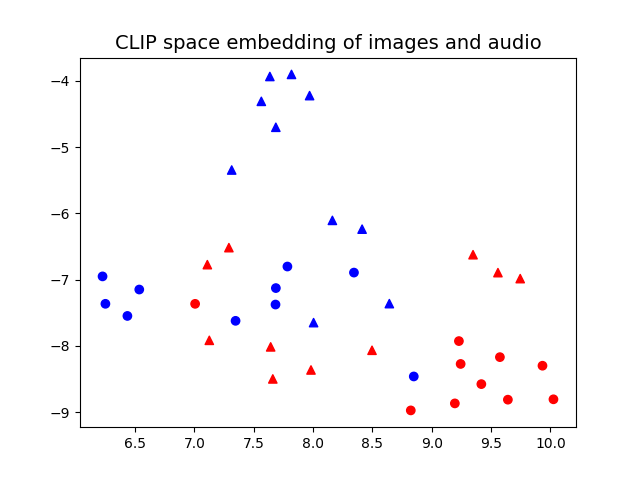}
         \end{subfigure}
     \caption{\textbf{Co-embedding of images and audio}. UMAP visualization of embeddings of audio and images from two distinct classes in one space. Audio used in Figure \ref{exp1a}, represented by circles, is co-embedded with images corresponding to those classes, represented by triangles. The resulting co-embedding manages to separate the two classes despite containing data in two distinct modes.}
     \label{exp2}
         \centering
         \begin{subfigure}[b]{0.6\textwidth}
         \centering
         \includegraphics[width=\textwidth]{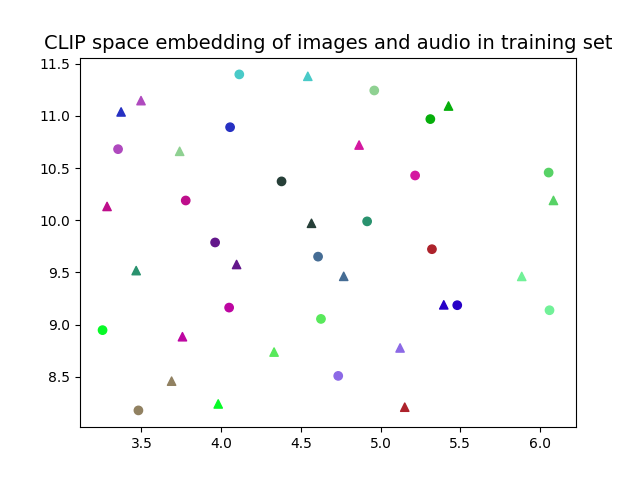}
         \end{subfigure}
     \caption{\textbf{Performance on training set}. UMAP visualization of co-embeddings of image-audio pairs randomly selected from the training set. Circles and triangles of the same color represent an image and audio file from the training data set.}
     \label{exp3}
\end{figure}

\subsection{Embedding Projections}

To demonstrate whether Embed Everything successfully learned a joint image-audio embedding, we embed several image-audio pairs and then project the embeddings into two dimensions using UMAP \cite{ umap}. We aim to determine whether the embedding space successfully separates data of different classes across modalities, while clustering data of the same class across modalities. 

As a baseline, we depict the VGGish embedding space for audio in Figure~\ref{exp1a}. In this case, we see that data of different classes is relatively separated.

Next, we examine how audio files are embedded in the CLIP embedding space, depicted in Figure \ref{exp1b}. We observe that audio data appears to be reasonably clustered by class, even when projected. This implies that the Embed Everything approach successfully captured some or all of the class separation information found in VGGish, and transferred that learning to the CLIP embedding. 

Finally, we explore how audio files and image files are clustered in the CLIP embedding space, shown in Figures \ref{exp2} and \ref{exp3}. In general, audio and image data of the same class appear to be colocated, while audio and image data of different classes are separated from each other. This strongly implies that the Embed Everything approach successfully learned a generalized image-audio embedding space.

\section{Discussion}
\label{others}

In this work, we present a novel, cost effective method of heterogeneous transfer learning, and show its efficacy by generating a useful audio-image co-embedding space.

Interestingly, even though we utilize CLIP as an image embedder, it actually generates a image-text co-embedding. As a result, we believe the embeddings generated by Embed Everything in our experiments actually live in a joint audio-image-text embedding space. In future work, we intend to explore how well text interacts with the audio in this space, especially since the transform layers were only trained on audio-image pairs.

As with other transfer learning approaches, the choice of the underlying models is key to successfully learning a task. In our case, we suspect that CLIP is a particularly good model for Embed Everything. Because CLIP learns a coembedding between text and images, the pretrained CLIP embedding space is likely to be sufficiently generalizable. Importantly, the Embed Everything approach cannot reasonably do better than the manifolds it is trained on, so it is important to select large models that have been trained on a significant amount of variable data. 

One limitation of the Embed Everything method is that its success is dependent on finding a dataset where a positive pairing is readily apparent. The Embed Everything system translates easily to image and audio since videos form a natural semantic link between the two modalities. However,
it may be difficult to find a publicly available dataset for any two arbitrary datatypes.

Additionally, the quality of the training dataset greatly impacts the accuracy of the model. Although VGGSound attempts to reduce ambient noise and present only visually apparent sources of audio, they maintain a 50\% threshold both for rejecting extraneous audio and for accepting correctly labeled class types. We believe this might cause difficulties since video clips containing extraneous noise or that are incorrectly identified ultimately interfere with the accuracy of our model.

Overall, we are excited to see that Embed Everything can make the task of creating joint embeddding spaces significantly easier. We look forward to expanding this work on other large models in other domains. 

\bibliographystyle{plain}
\bibliography{neurips_2021.bbl}
%%%%%%%%%%%%%%%%%%%%%%%%%%%%%%%%%%%%%%%%%%%%%%%%%%%%%%%%%%%%

\end{document}